%% file: arxiv.tex
\documentclass[10pt,journal,cspaper,compsoc]{IEEEtran}
\usepackage{lmodern}

\usepackage{times}
\usepackage{graphicx}
\usepackage{amsmath}
\usepackage{amssymb}

\usepackage{multirow}
\usepackage{color}
\usepackage[table]{xcolor}
\usepackage{overpic}
\usepackage{booktabs}
\usepackage{url}
\usepackage{algorithm}
\usepackage{algorithmic}
\usepackage[nocompress]{cite} 
\usepackage{ragged2e}

\usepackage{silence}
\usepackage[colorlinks]{hyperref}

\ifdefined \GramaCheck
  \newcommand{\CheckRmv}[1]{}
  \newcommand{\figref}[1]{Figure}
  \newcommand{\tabref}[1]{Table}
  \newcommand{\secref}[1]{Section}
  \newcommand{\appref}[1]{Appendix}
  \newcommand{\equref}[1]{Equation}
\else
  \newcommand{\CheckRmv}[1]{#1}
  \newcommand{\figref}[1]{Fig.~\ref{#1}}
  \newcommand{\tabref}[1]{Tab.~\ref{#1}}
  \newcommand{\appref}[1]{Appendix~\ref{#1}}
  \newcommand{\secref}[1]{Sec.~\ref{#1}}
  \newcommand{\equref}[1]{Eqn.~(\ref{#1})}
\fi

\def\eg{\emph{e.g.,~}}
\def\etc{\emph{etc}}
\def\etal{{\em et al.}}

\newcommand{\printfnsymbol}[1]{
  \textsuperscript{\@fnsymbol{#1}}
}
\graphicspath{{./figures/}{./CAM/}{./SAL/}}
\usepackage[normalem]{ulem}
\definecolor{awesome}{rgb}{1.0, 0.13, 0.32}
\definecolor{hyperref-green}{RGB}{0,150,0}
\definecolor{hyperref-blue}{RGB}{0,0,200}
\definecolor{hyperref-red}{RGB}{200,0,0}

\usepackage{comment}
\usepackage{wrapfig}
\usepackage{multirow} 
\usepackage{ifthen}
\usepackage{xstring}
\usepackage{overpic}
\usepackage{calc}
\usepackage{bm}
\usepackage{layouts}
\usepackage{pifont}
\usepackage{caption}

\graphicspath{{./fig/}}

\newcommand{\mathimg}[1]{\mathbf{#1}}

\def\supp{supplementary materials}

\newcommand{\tbdline}{\noindent\textcolor{red}{TBD }}
\newcommand{\tbd}[1][1]{ \newcount\tmp \tmp=0 \loop \advance\tmp by 1 \tbdline \ifnum\tmp<#1 \repeat }

\usepackage{xspace} 
\usepackage{mfirstuc} 
\newcommand{\defword}[2]{\def#1{\ifvmode\xmakefirstuc{#2}\else #2\fi \xspace}}
\defword\st{style transfer}

\defword\ist{IST}

\renewcommand{\paragraph}[1]{\noindent\textbf{#1}}

\renewcommand{\paragraph}[1]{\vspace{3pt}\noindent\textbf{#1}}

\begin{document}

\title{Interactive Style Transfer:~All is Your Palette}

\author{Zheng Lin, Zhao Zhang, Kang-Rui Zhang, Bo Ren, Ming-Ming Cheng
\IEEEcompsocitemizethanks{
\IEEEcompsocthanksitem Z. Lin, Z. Zhang, K.R. Zhang, B. Ren, and M.M. Cheng are with
TKLNDST, College of Computer Science, Nankai University,
Tianjin, 300350, China.
\IEEEcompsocthanksitem   Bo Ren is the corresponding author (rb@nankai.edu.cn).
}
}
\IEEEtitleabstractindextext{

\begin{abstract}
  \justifying
  Neural style transfer (NST) can create impressive artworks
  by transferring reference style to content image.
  Current image-to-image NST methods are short of fine-grained controls, 
  which are often demanded by artistic editing.
  To mitigate this limitation,
  we propose a drawing-like interactive style transfer (IST) method,
  by which users can interactively create a harmonious-style image.
  Our IST method can serve as a brush, dip style from anywhere, 
  and then paint to any region of the target content image.
  To determine the action scope,
  we formulate a fluid simulation algorithm, which takes styles as pigments around the position of brush interaction,
  and diffusion in style or content images according to the similarity maps.
  Our IST method expands the creative dimension of NST.
  By dipping and painting, 
  even employing one style image can produce thousands of eye-catching works.
  The demo video is available in supplementary files or in \url{http://mmcheng.net/ist}.
\end{abstract}

}

\input{main_parts/fig_introduction}

\maketitle
\IEEEdisplaynontitleabstractindextext
\IEEEpeerreviewmaketitle

\section{Introduction}
\label{sec:introduction}
\IEEEPARstart{N}{eural}  style transfer (NST) is a widely concerned artistic creation tool, 
which aims at transferring the artistic style from a reference image 
to a content image.
Many impressive NST methods have been proposed by exploring 
content retention~\cite{kotovenko2019content,cheng2021style}, 
robustness analysis~\cite{chiu2019understanding,wang2021rethinking}, 
ceiling of style types~\cite{huang2017AdaIN,chandran2021adaconv}, 
geometric changes~\cite{liu2021learning},
brushstroke simulation~\cite{kotovenko2021rethinking}, 
\etc.
Recently, several works have introduced the user's intention,
making NST can be manually adjusted in the degree of stylization
\cite{gatys2018factors}, 
semantic guidance~\cite{kurzman2019classbasedstyling,men2018texture}, 
or masked style splicing~\cite{virtusio2018interactive,stahl2019instanceist}.
Although these NST methods, 
even with global adjustment,
can create artworks, 
they can not provide fine-grained controls in artistic editing, 
and are not in line with natural drawing logic.

In this paper,
we propose a drawing-like interactive style transfer (IST) method, 
where users can dip the style from any region in the style images 
and then paint to any region of the target content image.
The scopes of dipping and painting are supported by our proposed fluid simulation algorithm,
making styles flow and blend like pigments in the content image according to the neural similarity maps.
Our IST method expands the creative dimension of traditional NST methods 
and brings greater imagination in creating harmonious-style artworks.
Some examples are shown in three subfigures of \figref{fig:introduction}, where, for example,
a) The user dips the whole style from style image and paints to content image with different interaction holding time.
The stylized scope in the content image first covers the clouds  and then fills the whole sky.
Compared to previous image-to-image NST,
our IST allows users to control the scope of stylization freely.
Therefore, users can transfer different styles to different regions of the content image.
b) Different areas also show separate style tendencies,
such as stroke direction and color concentration within a style image.
Our IST can freely dip the style on a specific region, then paint on the content image.
c) In drawing, the brush can be dipped in many pigments. 
Similarly, 
we can also dip multiple styles from one or multiple style images to blend in the target content region.
This interactive fusion can create thousands of styles by combination.
In addition to these examples, we show a complete artwork by our IST method in \figref{fig:teaser}.
More interesting creative scenes are shown in following sections and the supplementary demo video.

\input{main_parts/fig_teaser}

\section{Related Work}
\label{sec:related-work}

\subsection{Neural Style Transfer}

Neural style transfer (NST) 
\cite{chen2017stylebank,li2017WCT,TIP20_SP_NPR,huang2017AdaIN,lin2021drafting} 
aims to create novel artworks by transferring the artistic style 
from a reference image to a content image.
Making the transformation process more flexible and efficient has 
always been the focus of researchers.
\textbf{Single Style} methods~\cite{johnson2016perceptual,ulyanov2016texture}
learn a style-specific network by perceptual loss~\cite{johnson2016perceptual}, 
GAN~\cite{li2016precomputed}, normalization~\cite{ulyanov2017improved}, \etc. 
These earlier single style methods have to re-train new models for new styles.
\textbf{Multiple Style} methods
\cite{li2017diversified,chen2017stylebank,wang2017multimodal,zhang2018multi}
can transfer from multiple seen styles.
Such methods often save specific parameters for each style.   
However, in real-world applications, 
it is difficult to learn all styles in advance.
\textbf{Arbitrary Style} methods
\cite{huang2017AdaIN,li2017WCT,sheng2018avatar,park2019arbitrary,cheng2021style,wang2021rethinking,lin2021drafting}
can work on any given reference style image.
The online optimization methods 
\cite{gatys2016image,li2016combining,li2017laplacian} decouple 
content and style information of input images using a pre-trained model.
With iterative optimization,
the new artistic image imitates style image in the feature Gram matrix,
while maintaining similarity with the content image in feature representation.
Online optimization is time- and calculation-consuming.
Other methods tried to transfer arbitrary style by a single forward pass.
A popular practice
\cite{huang2017AdaIN,li2017WCT,cheng2021style,wang2021rethinking} 
directly aligns the statistics between the new artwork and the given style image 
in the instance feature space.
Some works~\cite{ghiasi2017exploring,shen2018neural,chandran2021adaconv} 
learn to predict dynamic parameters for any specified style.
Although such methods greatly improve the flexibility, 
the processing granularity is still full image and cannot meet the needs 
of all users.

\subsection{Style Transfer with Interactions}
Some related works have introduced user interactions,
in which the interactive part is mainly some ways to adjust the style.
Virtusio \etal~\cite{virtusio2021neural} allow users to control the ratio
of different sub-styles separated from a style image.
Akimoto \etal~\cite{akimoto2020soft} provide the fast soft color segmentation, 
which can extract and change the color distribution by users.
More works focus on the transferring with local areas.
But it is often handled by utilizing a mask, such as 
semantic mask~\cite{kurzman2019classbasedstyling} and 
instance mask~\cite{stahl2019instanceist}.
Many methods~\cite{gatys2018factors,alegre2020selfieart} adopt binary masks generated by users,
like brush area~\cite{dubey2019dataart} and 
object mask~\cite{virtusio2018interactive} by an interactive segmentation method~\cite{maninis2018dextr}.
In addition, there are some other operations, \eg
locally-retouching for style refinement~\cite{reimann2021interactive} and 
changing the semantic mask to generate an image with a new 
texture~\cite{men2018texture}.

However, these works based on adjusting the style parameters 
of the entire image are short of fine-grained controls.
The results from mask-based methods may look contradictory,
and the generation of masks may bring extra burdens. 
In this paper, we propose an IST method like the real drawing process with 
dipping and painting operations, which is not covered in the previous works.

\subsection{Fluid Simulation}

In computer graphics, fluid simulation is an active field, 
which studies how to simulate fluid in computers.
Controllable fluid simulation is essential for animators to achieve 
special effects and animations in 2D~\cite{witting1999computational} and 3D~\cite{stomakhin2017fluxed} scenes.
Many methods are proposed to calculate control force for the fluid,
such as blurred density gradient~\cite{fattal2004target},
proportional error of fluid shape between two key frames with velocity feedback~\cite{shi2005taming}, 
and optimal controllers~\cite{pan2017efficient}.
For image processing, some simplified forms of fluid motion equation, 
such as anisotropic diffusion equation~\cite{perona1990scale}, 
are used in denoising and edge detection.
In our work,
we introduce the fluid simulation algorithm for fine-grained controls with the scopes of interactions in IST.

\input{main_parts/fig_framework}

\section{Proposed Method}
\subsection{Framework Overview}
\label{sec:framework}
Our framework of \ist is inspired by traditional oil painting.
We treat the content image $\mathimg{C}$ as the painting canvas, and the style image set $\{\mathimg{S}^1~\dots~\mathimg{S}^n$\} as the painting palette.
The user dips the style on the style images by performing interactions, such as clicks and scribbles, 
and then similarly performs interactions to paint on the content image to transfer style with specific regions.

\figref{fig:framework} visually displays the interactive process with a single dipping and painting operation.
In the implementation of the algorithm, 
we input the content and style images into the backbone neural network to obtain their deep features in each layer, defined as $\mathimg{F}^i$,
calculate the similarity with the interaction location to obtain a global similarity map, defined as $\mathimg{G}$,~(see \secref{sec:similarity}),  
and then adopt the proposed similarity-based fluid simulation algorithm~(see \secref{sec:fluid}) to generate the penetration map $\mathimg{P}$ of interactions on the two images. 
We modify the classic AdaIN~\cite{huang2017AdaIN} method as the style transfer algorithm and propose the local style transfer~(see \secref{sec:transfer}) to apply the local style to the interactive region on the content image. 
Instead of simply splicing the stylized image through a binary mask,  
our \ist method will generate images with smooth transition by considering the soft style diffusion and blending.
Whether it is style dipping or painting, 
the degree of stylization for each pixel is a weight number.
If different styles are applied to the same position after multiple interactions, 
the styles are merged according to their fluid concentration rat.io
We also adopt the characteristics of oil painting to make subsequent styles more dominant in the result (see \secref{sec:multi-style}).

\subsection{Revisiting Neural Feature Similarity}
\label{sec:similarity}

As we know, neural network features show different levels of image characteristics~\cite{bau2017network}, from the low-level information, \eg, color and texture, to the high-level information, \eg, semantic.
For our framework, users take clicks or scribbles on the content and style images.
We also adopt these neural features to calculate the interaction radiation area.
In terms of concrete implementation, we use the widely used VGG~\cite{vgg} network as the backbone and utilize the first four layers. 
For each layer, we will calculate the corresponding similarity map by averaging the feature similarity between the user interaction and all pixels.
These user interactions $\mathimg{I}$ may contain only one pixel (click), or more (scribble).
Then we average the similarity maps in four layers into one $\overline{\mathimg{G}}$ as the basis of our fluid simulation algorithm.
We utilize the cosine similarity as the metric (shown in \equref{equ:cosine-similarity}).
The calculation process of the similarity map in the specific layer can be formulated as 
\begin{equation}
\label{equ:cosine-similarity}
\mathimg{G}_{m,n}=\frac{1}{\sum{\mathimg{I}}} \sum_{i,j}{\frac{\mathimg{F}_{m,n} \cdot \mathimg{F}_{i,j} \cdot \mathimg{I}_{i,j}}{\Vert\mathimg{F}_{m,n}\Vert \Vert\mathimg{F}_{i,j}\Vert}}, \mathimg{I}_{i,j} \in \{0,1\},
\end{equation}
where $\left(i,j\right)$ means the coordinates of all pixels and $\Vert\mathimg{F}_{i,j}\Vert$ means the 2-norm value in position $\left(i,j\right)$. 
In \figref{fig:similarity}, we illustrate the similarity maps of four layers according to the click.
From $\mathimg{G}^1$ to $\mathimg{G}^4$, 
the similar regions transit from similar colors to similar semantics.
On the whole, 
the deeper the features are, the information concerned becomes more complex and the similar area is smaller.
\input{main_parts/fig_similarity}

\vspace{-10pt}
\subsection{Similarity-Based Fluid Simulation}  
\label{sec:fluid}
\begin{figure}[t]
  \centering
  \includegraphics[width=1.0\linewidth]{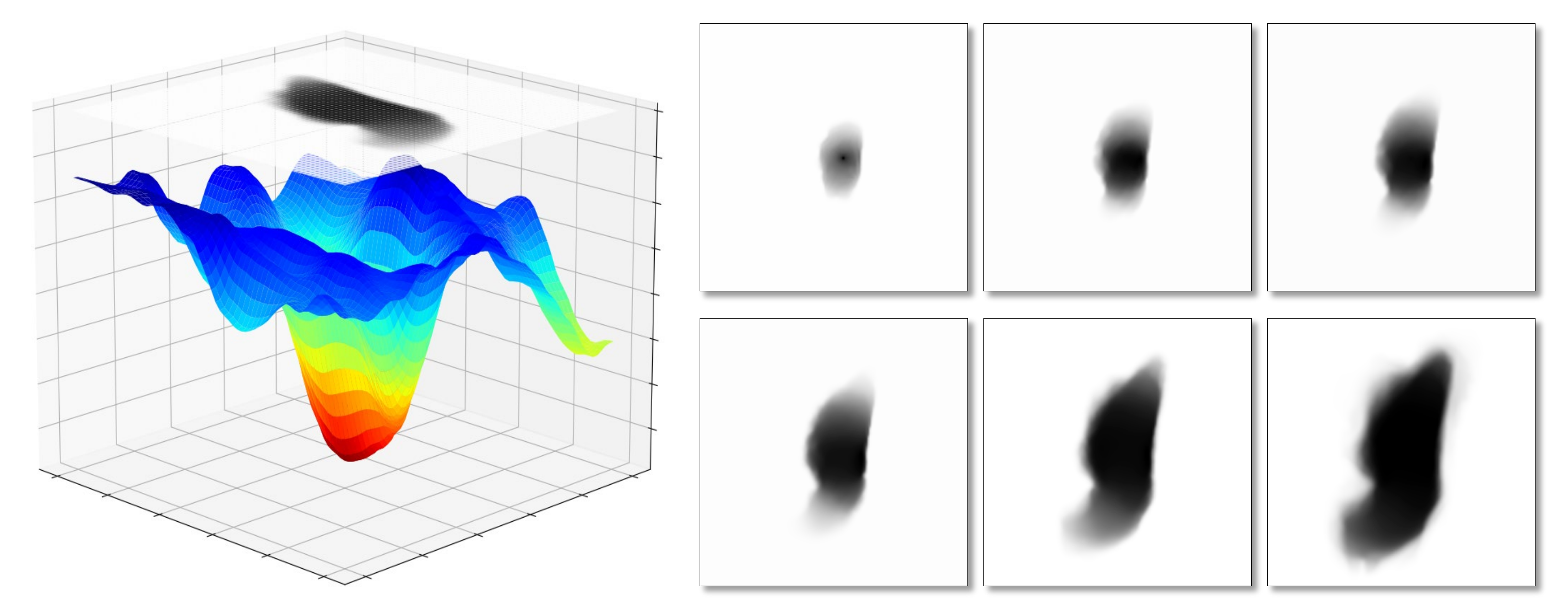}
  \caption{Illustration for similarity-based fluid simulation. 
  The left part shows that our fluid diffusion is according to the similarity map,
  shown as the curved surface.
  The right part shows penetration maps in the diffusion process.
  } 
  \label{fig:fluid}
  \vspace{-15pt}
\end{figure}

The global similarity maps  measure the correlation between the interaction position and the pixels of the whole picture. 
However, in the interactive process, users often hope to obtain the style from or transfer the style to the areas around the interaction.
More importantly, users often need to control the scope of painting, similar to the brush hardness or concentration in painting.
Therefore, we introduce an algorithm of similarity-based fluid simulation to generate the influence area of user interactions.
Since the diffusion and absorption of ink can be regarded as an opposite process, we use the same algorithm for dipping and painting style.
\figref{fig:fluid} shows the concept of our algorithm.
Generally speaking, 
the similarity is regarded as the resistance in fluid diffusion.
The higher the similarity, the smaller the resistance. 
From the user interaction position, the fluid concentration will gradually diffuse outwards.

For multiple styles, in contrast to generating hard areas consisting of Boolean values 0 and 1, our purpose is to generate a smooth transition on the border.
Pixels from where the user directly interacts with  have a higher concentration of fluids, which is diffused to surrounding pixels having a lower concentration over time. 
We therefore use a fluid-simulation like algorithm to achieve our goal.
Specifically,  
a penetration function $\mathimg{P}\left(x,y,t\right)$ represents a distribution of fluid concentration.
And the similarity-based diffusion coefficient function $\mathimg{D}\left(x,y\right)$ represents a parameter to  control the velocity of fluid diffusion.
The 2D fluid diffusion equation can be written as 
\begin{equation}
\label{equ:diff}
\begin{aligned}
\frac{\partial \mathimg{ P}{\left(x,y,t\right)}}{\partial t}
&= \nabla \cdot\left( \mathimg{ D}{\left(x,y\right)} \nabla  \mathimg{ P}{\left(x,y,t\right)} \right),
\end{aligned}
\end{equation}
in interval $(x,y)\in[0,1]\times[0,1] $ with the initial condition 
$\mathimg{ P}{\left(x,y,0\right)} = \mathimg{ I}{\left(x,y\right)}$
where $\mathimg{ I}{\left(x,y\right)}$ represents our interactive position in continuous space,
and the Neumann boundary condition, $ \partial\mathimg{P}{\left(x,y,t\right)}/\partial n = 0$ on the boundary where $n$ represents the normal direction to the boundary.
$\nabla \cdot$ and $\nabla$ represent the divergence  operator and gradient operator respectively.
$\mathimg{P}{\left(x,y,t\right)}$ is a continuous function of position $\left(x,y\right)$ and time $t$.
$\mathimg{D}{\left(x,y\right)}$ and $\mathimg{ I}{\left(x,y\right)}$ are  continuous functions of position $\left(x,y\right)$.
It needs to be emphasized that a similarity-based $\mathimg{D}{\left(x,y\right)}$ in \equref{equ:diff} makes the fluid concentration diffuse more on the area with higher similarity.
In contrast, 
the standard isotropic diffusion model diffuses concentration uniformly along every direction,
which shows less controllability. 
For the convenience of discretization of \equref{equ:diff}, we write it in the form of components:
\begin{equation}
\label{equ:substitute}
\begin{aligned}
\frac{\partial \mathimg{P}{\left(x,y,t\right)}}{\partial t}
= &\frac{\partial }{\partial x} \left(\mathimg{D}{\left(x,y\right)}
\frac{\partial }{\partial x}  \mathimg{P}{\left(x,y,t\right)}\right)\\
+&\frac{\partial}{\partial y} \left(\mathimg{D}{\left(x,y\right)}\frac{\partial }{\partial y}  \mathimg{P}{\left(x,y,t\right)}\right).
\end{aligned}
\end{equation}

\paragraph{Discretization scheme.}
In order to make the calculation domain of the discretized diffusion equation just on the image we care about, we follow the discrete domain of the image, consisting of $N_i \times N_j$ pixels. 
We write $\mathimg{P}_{i,j}^n$ for $\mathimg{ P}{\left(i \Delta h_i,j \Delta h_j,n \Delta t\right)}$, $\mathimg{D}_{i,j}$ for $\mathimg{D}{\left(i \Delta h_i,j \Delta h_j\right)}$, where $\Delta h_i = 1/N_i $, $ \Delta h_j = 1/N_j$, and $\Delta t$ is a given time step.
Here, we could define penetration map $\mathimg{ P}^n$ and diffusion coefficient map $\mathimg{ D}$ as two matrices consisting of $\mathimg{P}_{i,j}^n$ and $\mathimg{D}_{i,j}$, respectively.
An implicit backward difference in time method~\cite{dehghan1999fully} is used to solve \equref{equ:substitute} stably and quickly.
The discretization scheme of \equref{equ:substitute} is formulated as 
\begin{equation}
\label{equ:discDiff}
\begin{aligned}
\frac{ \mathimg{ P}_{i, j}^{n+1}-\mathimg{ P}_{i,j}^{n}}{\Delta t}
&=\Big(\mathimg{ D}_{i+1/2,j} \left(\mathimg{ P}_{i+1, j}^{n+1}-\mathimg{ P}_{i, j}^{n+1}\right)\\
&-\mathimg{ D}_{i-1/2,j} \left(\mathimg{ P}_{i, j}^{n+1}-\mathimg{ P}_{i-1, j}^{n+1}\right)\Big)/\Delta x^2 \\
&+\Big(\mathimg{ D}_{i,j+1/2} \left(\mathimg{ P}_{i, j+1}^{n+1}-\mathimg{ P}_{i, j}^{n+1}\right)\\
&-\mathimg{ D}_{i,j-1/2} \left(\mathimg{ P}_{i, j}^{n+1}-\mathimg{ P}_{i, j-1}^{n+1}\right)\Big)/\Delta y^2.
\end{aligned}
\end{equation}
$\mathimg{ P}_{i,j}^{0}$ is already known according to the initial condition.
The calculation of $\mathimg{D}$ and the diffusion coefficient value on  staggered grid, such as $\mathimg{ D}_{i+1/2,j}$, will be describe in detail later.
\equref{equ:discDiff},
with discretized initial condition and boundary condition,
that traverses all $i$ and $j$, is a large sparse linear system. Its  coefficient matrix is symmetric and positive definite.
We could use a standard conjugate gradient method~\cite{bolz2003sparse} to solve this linear system quickly. 

\paragraph{Diffusion coefficient.}
The calculation process of diffusion coefficient is formulated as follows: 
\begin{equation}
\label{equ:diffCoef}
\begin{aligned}
\mathimg{D}_{i,j} = v e^{- r\left(1-\mathimg{G}_{i,j}\right)},
\end{aligned}
\end{equation}
which is dependent on similarity nonlinearly. 
The main benefit of using this non-linear function is to diffuse nearly no concentration where similarity is relatively low to make a visible  difference with isotropic diffusion.
$v$ is the diffusion velocity to control global diffusion effects. 
$r$ is the resistance coefficient to control local diffusion effects. The higher $r$ is, the distribution of $e^{- r\left(1-\mathimg{G}_{i,j}\right)}$ becomes sharper,
diffusion area may shrink,
and vice versa,     
the distribution gets flatter,
and diffusion area may expand more.
An interpolation method is shown below to calculate the diffusion coefficient on our staggered grid: 
\begin{equation}
  \label{equ:inte}
  \begin{aligned}
  \mathimg{D}_{i+1/2,j} = \min\left(\mathimg{D}_{i,j},\mathimg{D}_{i+1,j}\right).
  \end{aligned}
\end{equation}
Other methods, such as the linear interpolation and the maximum function,
could achieve the similar effect except for a slight difference to the diffusion on the border. 
All commutative functions preserve the conservation of concentration. 

\paragraph{Automatic termination.}
These diffusion processes will continue all the time.
Our \ist method provides a manual way so that users can stop the diffusion at the specific time when they are satisfied. 
But when the users want to interact more conveniently,
we also provide an automatic mode.
When the rate of variation in concentration is less than a certain value $\epsilon$, which is formulated as 
\begin{equation}
\frac{\sum_{i,j}{\vert\mathimg{P}_{i,j}^{n+1}-\mathimg{P}_{i,j}^{n}\vert}} {\sum_{i,j}{\mathimg{P}_{i,j}^{n}}} \le \epsilon,
\end{equation}
the fluid diffusion is stopped.
With the automatic way, users can create artworks efficiently.

\subsection{Local Style Transfer}
\label{sec:transfer}
At present, most NST methods extract the style from the whole picture and transfer it to another picture.
The research on partial style extraction is less studied.
Since the classic method of NST, AdaIN~\cite{huang2017AdaIN}, focuses on the mean and variance of neural features from content and style images, 
and these two values can be calculated locally, 
we modify the method to adapt to the local transfer in our \ist.

\paragraph{Revisiting AdaIN.}{
The AdaIN method is based on the deduction that the style of the image is determined by the mean and variance of the deep features.
This method first puts the corresponding content $\mathimg{C}$ and style $\mathimg{S}$ into the backbone network to get the deep features, $\mathimg{F}^c$ and $\mathimg{F}^s$.
Then the content features will be normalized according to its mean and variance, 
and finally restored using the mean and variance of the style features, 
so that the generated features $\mathimg{A}$ contain the content of the content image and the style of the style image.
The calculation process is as follows:
\begin{equation}
\label{equ:adain}
\mathimg{A}=\sigma\left(\mathimg{F}^s\right)\left(\frac{\mathimg{F}^c-\mu\left(\mathimg{F}^c\right)}{\sigma\left(\mathimg{F}^c\right)}\right)+{\mu\left(\mathimg{F}^s\right)},
\end{equation}
where $\mu$ and $\sigma$ represent the function to calculate the channel-wise mean and standard deviation, respectively.
The generated features will be fed into a decoder network and be restored to the transferred image $\mathimg{T}=\mathtt{Decoder}(\mathimg{A})$.}

\paragraph{Modified local AdaIN.}
We modify the original AdaIN to carry out local style transfer.
For style extraction, we use the normalized penetration map $\mathimg{P}$ to calculate the weighted mean and variance of style features.
It can be formulated as 
\begin{equation}
\label{equ:local-adain}
\mathimg{A}=\tilde{\sigma}\left(\mathimg{F}^s,\mathimg{P}^s\right)\left(\frac{\mathimg{F}^c-\mu\left(\mathimg{F}^c\right)}{\sigma\left(\mathimg{F}^c\right)}\right)+{\tilde{\mu}\left(\mathimg{F}^s,\mathimg{P}^s\right)},
\end{equation}
where $\tilde{\mu}$  and $\tilde{\sigma}$ represent the function to calculate the weighted mean and standard deviation as follows:
\begin{equation}
\label{equ:weighted-mean}
\tilde{\mu}\left(\mathimg{F},\mathimg{P}\right)= \frac{ \sum_{i,j} \mathimg{P}_{i,j} \cdot \mathimg{F}_{i,j}}{\sum_{i,j}{\mathimg{P}_{i,j}}},
\end{equation}
\begin{equation}
\label{equ:weighted-std}
\tilde{\sigma}\left(\mathimg{F},\mathimg{P}\right)=   \sqrt{ \frac{ \sum_{i,j} \mathimg{P}_{i,j} \cdot  \left(\mathimg{F}_{i,j}-\tilde{\mu}\left(\mathimg{F},\mathimg{P}\right)\right)^2 }{\sum_{i,j}{\mathimg{P}_{i,j}}}}.
\end{equation}
Then we will merge the transferred feature and the original content features to get $\mathimg{M}$ and feed it into the decoder network to get the transferred image $\mathimg{T}$:
\begin{equation}
\label{equ:merge-decoder}
\mathimg{T}=\mathtt{Decoder}(\mathimg{M}),\mathimg{M}= (1-\mathimg{P}^c) \cdot \mathimg{F}^c + \mathimg{P}^c \cdot \mathimg{A}.
\end{equation}
Because the neural network will cause partial distortion of the image, even when the values in $\mathimg{P}$ is all equal to 0. 
So when we generate the final output $\mathimg{O}$, we mix the original image and the transferred image $\mathimg{T}$ for the low-penetration area:
\begin{equation}
\label{equ:ori-merge}
\mathimg{O}=\mathimg{C} \cdot \phi\left( \mathimg{R}\right) + \mathimg{T} \cdot \left(1-\phi\left(\mathimg{R}\right)\right),
\end{equation}
\begin{equation}
\phi\left(\mathimg{R}\right)=\max\left(\frac{\mathimg{R}-\alpha}{1-\alpha},0\right),
\end{equation}
where $\mathimg{R}$ is the content retention map and is equal to $1-\mathimg{P}^c$ when these is only one interaction and $\phi$ is to calculate the specific proportion of the original content image with $\alpha$ as an factor. 

These above show the case where there is only one style and content image, 
such as the cyan scribble `\textcolor[rgb]{0.239,0.737,0.792}{$\bm{\wr}$}' in \figref{fig:introduction} (c).
Suppose we want to fuse some areas of multiple style images to create a new style, such as the purple scribble `\textcolor[rgb]{0.709,0.380,0.627}{$\bm{\wr}$}' in \figref{fig:introduction} (c).
As long as these pixels and corresponding concentrations are regarded as a whole, 
the corresponding weighted mean and standard deviation can be obtained to represent the new style.

\vspace{-8pt}
\subsection{Multi Styles Mixture}
\label{sec:multi-style}
\ist often applies multiple styles to different areas of the picture. 
These styles may come from one picture or multiple pictures,
and the same style may also correspond to many regions.
It is worth considering how to mix when the same pixel position is affected by multiple styles.
We adopt a treatment method similar to oil painting. 
When the new style and interaction occur,
the subsequent style will be mixed with the previous ones, 
and the subsequent one will play a dominant role. 
From the perspective of the fluid, 
the previous concentration based on the fluid simulation algorithm will be retained to a certain extent at these locations,
while the subsequent diffusion results will dominate.
In terms of the specific implementation,
we adopt an iterative calculation. 
The current mixed feature $\mathimg{M}^{[n]}$ is generated based on the transferred features $\mathimg{A}^{[n]}$ by the current style and the previously saved mixed features $\mathimg{M}^{[n-1]}$ according to the penetration map on content:
\begin{equation}
\label{equ:feat-merge}
\mathimg{M}^{[n]}=(1-\mathimg{P}^{[n]}) \cdot \mathimg{M}^{[n-1]} + \mathimg{P}^{[n]} \cdot \mathimg{A}^{[n]},~~\mathimg{M}^{[0]}=\mathimg{F}^c.
\end{equation}
As shown above, the initial mixed feature is the same as $\mathimg{F}^c$,
and the feature will gradually integrate more and more features of different styles.
The $\mathimg{P}$  in \equref{equ:feat-merge} means the penetration map in content image and the index $[n]$ is used to indicate interaction sequence.
With the increase in interactions on the content image, the content retention map is also gradually affected, which is expressed as follows:
\begin{equation}
\mathimg{R}^{[n]}=\mathimg{R}^{[n-1]} \cdot \left(1-\mathimg{P}^{[n]}\right),~~\mathimg{R}^{[0]}=\mathimg{1}.
\end{equation}

\subsection{Interactive Process and User Interface}
For \ist, the basic operations consist of two parts, \eg, dipping on style and painting on content.
The actual interactive interface and operation video will be shown in the \supp. 
We will introduce these two operations with the user interface (UI) prototype shown in \figref{fig:ui}.
The UI is mainly divided into two parts: the left content canvas and the right style palette.
The style palette may contain multiple style images, such as four in the prototype.
Users can provide points `$\bullet$' and scribbles `$\bm{\wr}$' for dipping and painting.
We also provide a convenient operation to dip or paint on the whole image, 
shown as the star symbol `$\bigstar$' on the upper left corner.
This can be implemented by a special action, such as the middle button of the mouse.

\input{main_parts/fig_ui}

\input{main_parts/fig_abl-fluid}
\input{main_parts/fig_iis}
\paragraph{Dipping on style.}{
Different pictures will be different in style.
Even in the same picture, the styles in different positions vary. 
For example, the style of a sketch, 
the stroke and the hardness are all diverse in different positions.
The user clicks or scribbles on the specific area on one of the style images.
It will calculate a penetration map based on the automatic fluid simulation algorithm according to the similarity maps.
The modifed AdaIN will extract the local feature and get the transferred deep features.
It is worth mentioning that if the user wants to select two areas in two style images and merge them as a new style, 
they can take a specific action to combine the two interactions,
such as holding the `Control Key' in computer and performing interactions on multiple style images,
as shown in \figref{fig:ui}~\ding{195}.  
}

\paragraph{Painting on content.}{
For the extracted local style,
the user can choose to add one or more interactions to the content image.
Different from dipping on style, 
we can choose manual or automatic generation for the penetration map.
For automatic way, it is similar to style dipping. 
After performing interactions, 
the fluid simulation algorithm will calculate a stable state as our penetration map.
For the manual way, 
users can take an additional action, 
such as pressing the `Shift Key' in computer before the interaction.
After the interaction, 
our method will generate a corresponding transferred result according to the process of fluid diffusion, 
which looks like a style transfer that continuously diffuses outward from the interaction. 
When the user is satisfied at a certain moment, release the corresponding action to stop.
For the click interaction in computer,
users can directly control the style diffusion process by pressing and releasing the right mouse button.
}

\section{Experiments}
\subsection{Method Analysis}
\paragraph{Implementation details.}
In terms of \ist, the $\alpha$ is set to 0.7, the $\epsilon$ is set to 0.01, and the default $v$ and $r$ are set to 1.0 and 10.
In terms of the specific implementation,
the training of AdaIN is the same as that in \cite{huang2017AdaIN}.
We adopt the VGG-19 as our backbone neural network.
The implementation of the neural network adopts the PyTorch framework.
The fluid simulation algorithm is based on Compute Unified Device Architecture (CUDA).
The UI for demo video is implemented based on Matplotlib.
All experiments are run on a single NVIDIA Titan XP GPU.

\paragraph{Fluid simulation analysis.}
We ablate the parameters $v$ and $r$ in our similarity-based fluid simulation algorithm by showing the animation sequence in \figref{fig:abl-fluid}.
The diffusion velocity is reflected in the left part. 
As it increases, it can be seen that the diffusion speed of the fluid increases significantly. 
In addition, due to the increase in diffusion velocity, the difference between the two steps becomes larger, affecting the automatic results.
From the right part, we change the resistance coefficient. 
The effect of the resistance coefficient is negatively related to the diffusion velocity to a certain extent. 
But the different is that as the resistance coefficient increases, 
the more difficult it is to diffuse on the border, 
the more obvious the border will become. 
Our fluid simulation method can produce real-time animation, which reach 174 FPS on 512$\times$512 images.
Adding the style transfer method, it can achieve 53 FPS.

\paragraph{Comparing to mask-based transfer idea.}
An intuitive idea for transfering style to local regions is eliminating irrelevant background with a segmented mask, 
which can be obtained from interactive segmentation~\cite{sofiiuk2020fbrs,lin2020interactive} or other automatic segmentation technology.
However, two key factors make the problem it solves is fundamentally different from ours.
1) The process of selecting the scope by interactive segmentation is far from the logic of human drawing. 
Interactive segmentation methods focus on the boundary and intend to segment the whole entity.
Unlike the feel of using a painting brush, 
the user cannot control the scope of action as much as they want.
As shown in the left part of \figref{fig:iis},
when using the interactive segmentation method~\cite{lin2020interactive} to segment a part of the sky, 
the interaction process becomes very tedious. 
Besides, since there is no clear boundary inside the sky, the edges of the area are jagged due to uncertainty.
Our method brings a drawing-like interactive style transfer idea, where users can dynamically dip local style and paint to arbitrary regions with ﬂexible scope and strength.
With one click, the style spreads naturally from the brush. 
This process is more conform to human drawing logic.
2) Segmented masks obtained from interactive/automatic segmentation technology suffer from the hard boundary during style transferring, and are difficult to achieve the natural transition and blending of styles.
As shown in the right part of \figref{fig:iis},
when we transfer styles to the pre-segmented regions of the dress, the boundary of the two styles is obviously rigid and unnatural.
Using our method,
the styles transitions smoothly and blend naturally with fluid simulation.
The two factors make the idea cannot serve as an alternative baseline and hard to be compared quantitatively.

\input{main_parts/fig_dip}
\input{main_parts/fig_paint}

\vspace{-6pt}
\subsection{Qualitative Results \& Discussions}

\paragraph{What can dipping create?}
In our IST method,
we first dip style by clicking or scribbling on style images.
Compared with transferring the style of the whole reference image,
our dipping process can create diverse works that meet the user's intentions.
We show some examples in \figref{fig:dip}.
In the first row,
the content picture is the flower with a butterfly,
and the style picture is a sketch-style portrait painting.
Employing AdaIN~\cite{huang2017AdaIN},
users cannot control the detailed style of the artwork;
while, using IST,
the artworks can present different stroke directions and color concentrations.
The second row shows more complex scenes.
Users dip and merge styles from one or multiple images and create dazzling artworks with limited reference images.
The figure is not for aesthetics, but for showing that our method can ﬁne-grainedly extract styles, such as different colors and stroke directions.

\paragraph{What can painting create?}
After dipping the style,
we can paint by clicking or scribbling on content images.
We illustrate some cases in \figref{fig:paint}.
The first row contains a content image of buildings and a style image.
We can dip style and paint it on different regions of the building.
This operation can change the area that the user wants to change, rather than apply to the whole image.
The second row shows the style flow from the interaction position with the increase of holding time,
and gradually turning the grassland into a lake.
This style-aware fluid diffusion process is as natural and controllable as liquid pigment.
The third row shows an interesting scene.
We dip four different styles from two style images and paint them onto three flowers and the background.
We exchange styles between them to produce a variety of artworks.
Because of fine-grained control, our method can freely express the creativity of users.

\vspace{2pt}
\paragraph{Artworks freely created by users.}
We have invited some people to enjoy our IST method.
Some of these artworks are shown in \figref{fig:result_all},
which are created by their interactions and integrate a variety of styles. 
The main factors affecting aesthetics are the creator’s creativity and carefulness.

\vspace{-20pt}
\section{Conclusion}
\vspace{-5pt}
In this paper,
we bring a new creative
dimension to the neural style transfer (NST).
It breaks the image-to-image transferring limitation and formulates the NST as a drawing process.
Users can dip style from one or multiple style images and then paint it to any region of the content image.
We call the creation process interactive style transfer (IST).
In our IST method,
we employ a fluid simulation algorithm to simulate the style dipping and painting process with brush.
Its working scope is determined by the neural similarity from interaction position and holding time.
We create interesting artworks using our IST method in the paper and show some creation processes in the supplementary video.
\input{main_parts/fig_result}

{\small
\bibliographystyle{ieee}
\bibliography{arxiv}
}

\vfill

\end{document}

%% file: main_parts/fig_introduction.tex
\def\IEEEcompsocdiamondline{

{\vrule depth 0pt height 0.5pt width 4cm\nobreak\hspace{7.5pt}\nobreak
\raisebox{-3.5pt}{\fontfamily{pzd}\fontencoding{U}\fontseries{m}\fontshape{n}\fontsize{11}{12}\selectfont\char70}\nobreak
\hspace{7.5pt}\nobreak\vrule depth 0pt height 0.5pt width 4cm\relax}

\setcounter{figure}{0}
\centering
\vspace{5pt}
\begin{overpic}[width=0.96\linewidth]{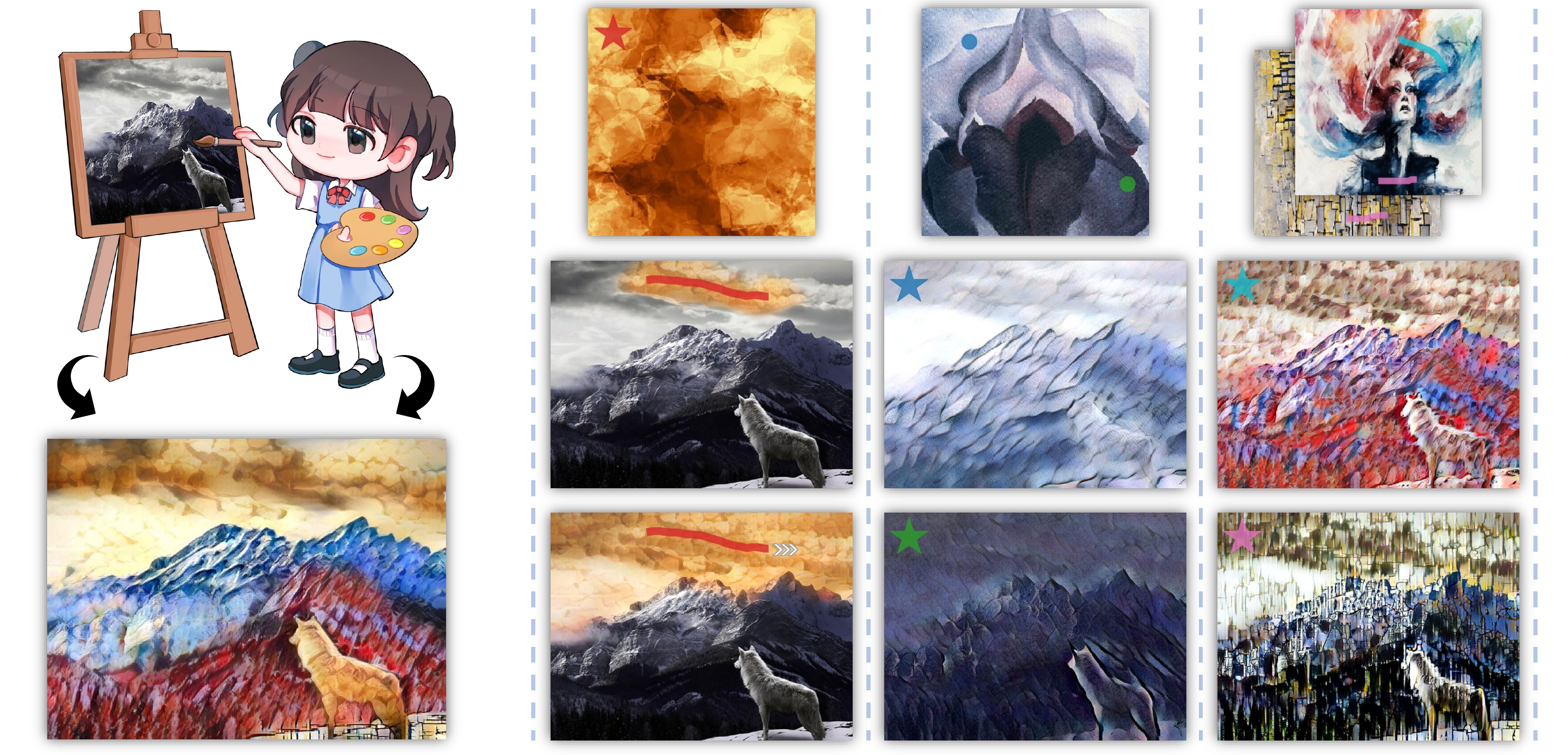} 
\put(11.5,21.3){\emph{\large{artwork}}}
\put(7.1,33.1){  \rotatebox{7.618}{ \emph{\footnotesize{content}}}  }
\put(30.5,36.3){  \rotatebox{90}{ \emph{\large{styles}}}  }
\put(30.5,12.3){  \rotatebox{90}{ \emph{\large{results}}}  }
\put(34.6,45.7){\normalsize{(a)}}
\put(55.9,45.7){\normalsize{(b)}}
\put(77.2,45.7){\normalsize{(c)}}
\end{overpic}
\vspace{-5pt}
\captionsetup{type=figure}
\captionof{figure}{
  The proposed interactive style transfer (IST) method allows users to transfer styles like drawing process.
  Users can dip styles from any position in multiple style images and create attractive artworks on the content image.
  With interactions, some interesting cases are shown above.
  (a) Same style image, different scopes of painting;
  (b) Same style image, different dipping regions to create abundant styles;
  (c) Dip and merge various styles in brush on either one or more images.
  The point `$\bullet$' and scribble `$\bm{\wr}$' denote user interactions,
  and the star symbol `$\bigstar$' indicates 
  the dipping or painting operations on the whole image.
  The demo video shows more cases.
}
\label{fig:introduction}
\vspace{-30pt}
}

%% file: main_parts/fig_teaser.tex
\begin{figure}[t]
  \centering
  \begin{overpic}[width=1.0\linewidth]{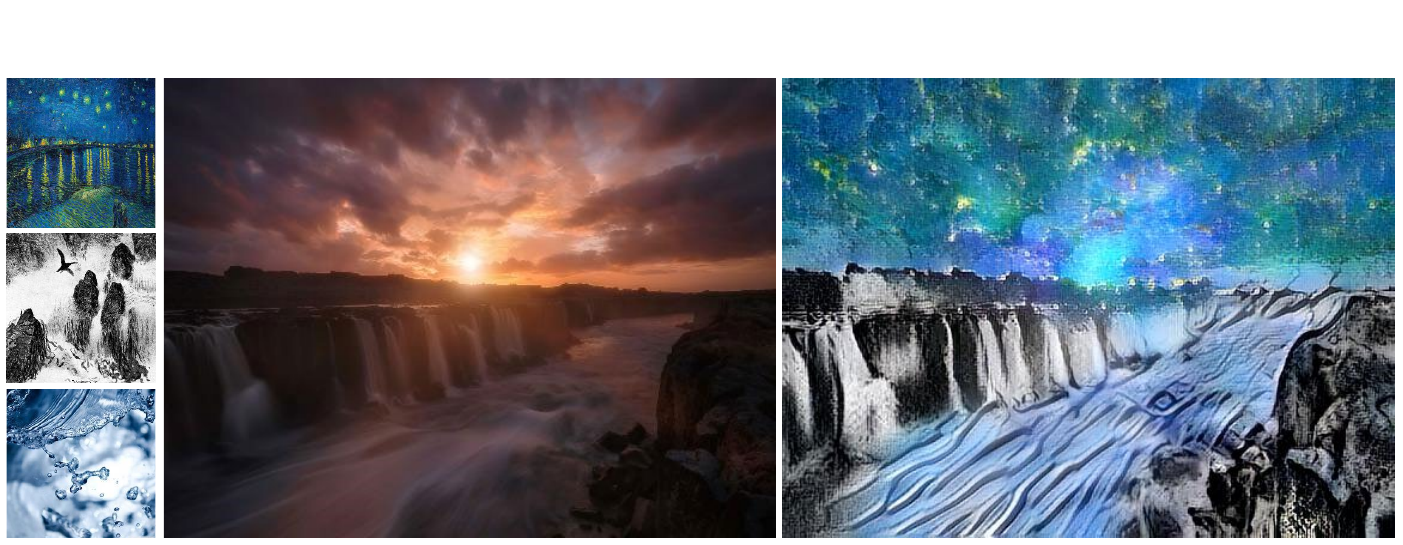}
      \put(1.4,35.0){Style}
      \put(27.0,35.0){Content}
      \put(73.0,35.0){Result}
  \end{overpic}
  \caption{
      An artwork created by our IST method.
      We omit the interaction traces in the above images.
      By dipping and painting operations,
      we decorate the sky with stars, the mountains with ink, and the river with light.
      } 
\label{fig:teaser}
\vspace{-12pt}
\end{figure}

%% file: main_parts/fig_framework.tex
\begin{figure*}[t]
  \centering
	\includegraphics[width=0.95\linewidth]{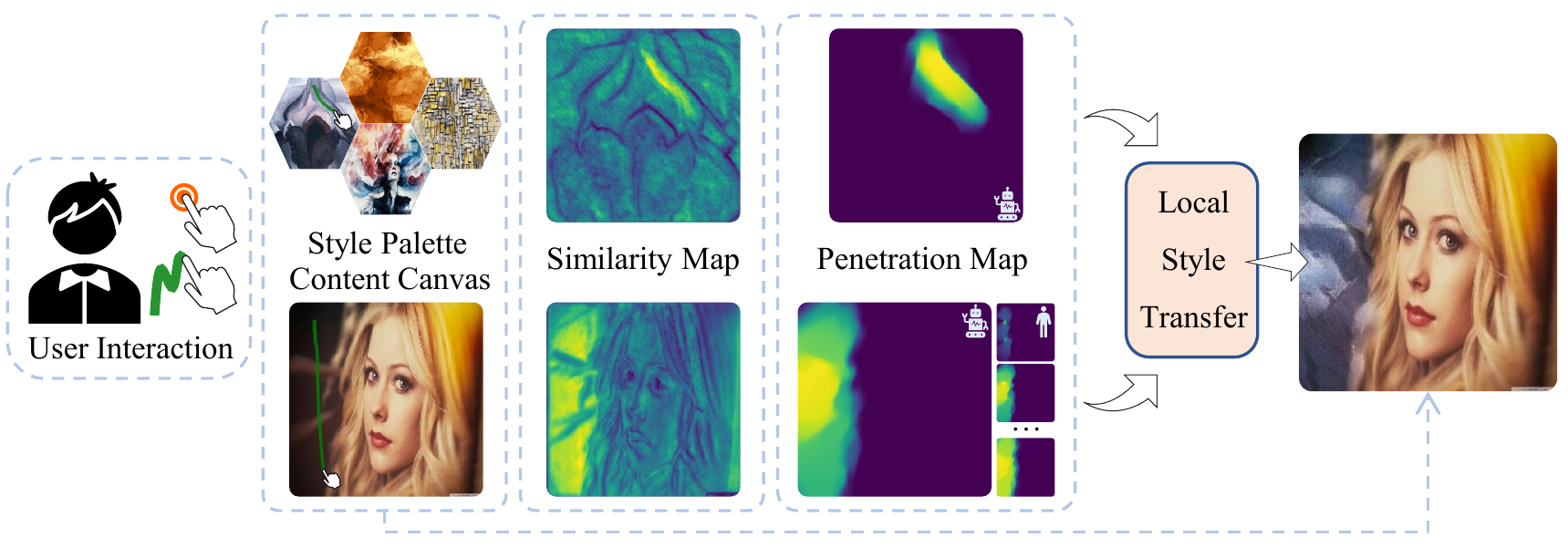}
	\vspace{-8pt}
	\caption{The framework of an interactive process. 
  User take interactions on style and content images.
  The similarity map will be calculated and used in the 
  fluid simulation algorithm and generate the penetration map. 
  (The automatic way for style images and an additional manual way for content images.)
  Then the local style transfer method will be adopted to generate the result.
  } 
  \label{fig:framework}
  \vspace{-12pt}
\end{figure*}

%% file: main_parts/fig_similarity.tex
\begin{figure}[tbp]
    \centering
    \def\wideratio{0.30}
    \newcommand{\mylabel}[3]{\put(#1,#2){   \textcolor{yellow}{  \large{\textbf{#3}}   }}}
    \newcommand{\tmpline}[2]{\begin{overpic}[width=.30\linewidth]{similarity/#1} \mylabel{2}{7}{#2}\end{overpic}\hspace{-1.5pt}}
    
    \tmpline{sim_000_sailboat_img.jpg}{{$\mathimg{I}$}}
    \tmpline{sim_000_sailboat_sim-1.jpg}{$\mathimg{G}^1$}
    \tmpline{sim_000_sailboat_sim-2.jpg}{$\mathimg{G}^2$}
    
    \vspace{1pt}
    
    \tmpline{sim_000_sailboat_sim-5.jpg}{$\overline{\mathimg{G}}$}
    \tmpline{sim_000_sailboat_sim-3.jpg}{$\mathimg{G}^3$}
    \tmpline{sim_000_sailboat_sim-4.jpg}{$\mathimg{G}^4$}
    \caption{Neural similarity maps based on the interaction location. 
    $\mathimg{G}^i$ shows the similarity maps of features in four layers based on the click interaction $\mathimg{I}$.
    The average of all is shown as $\overline{\mathimg{G}}$.} 
    \label{fig:similarity}
    \vspace{-15pt}
\end{figure}

%% file: main_parts/fig_ui.tex
\begin{figure}[t]
    \centering
    \begin{overpic}[width=1.0\linewidth]{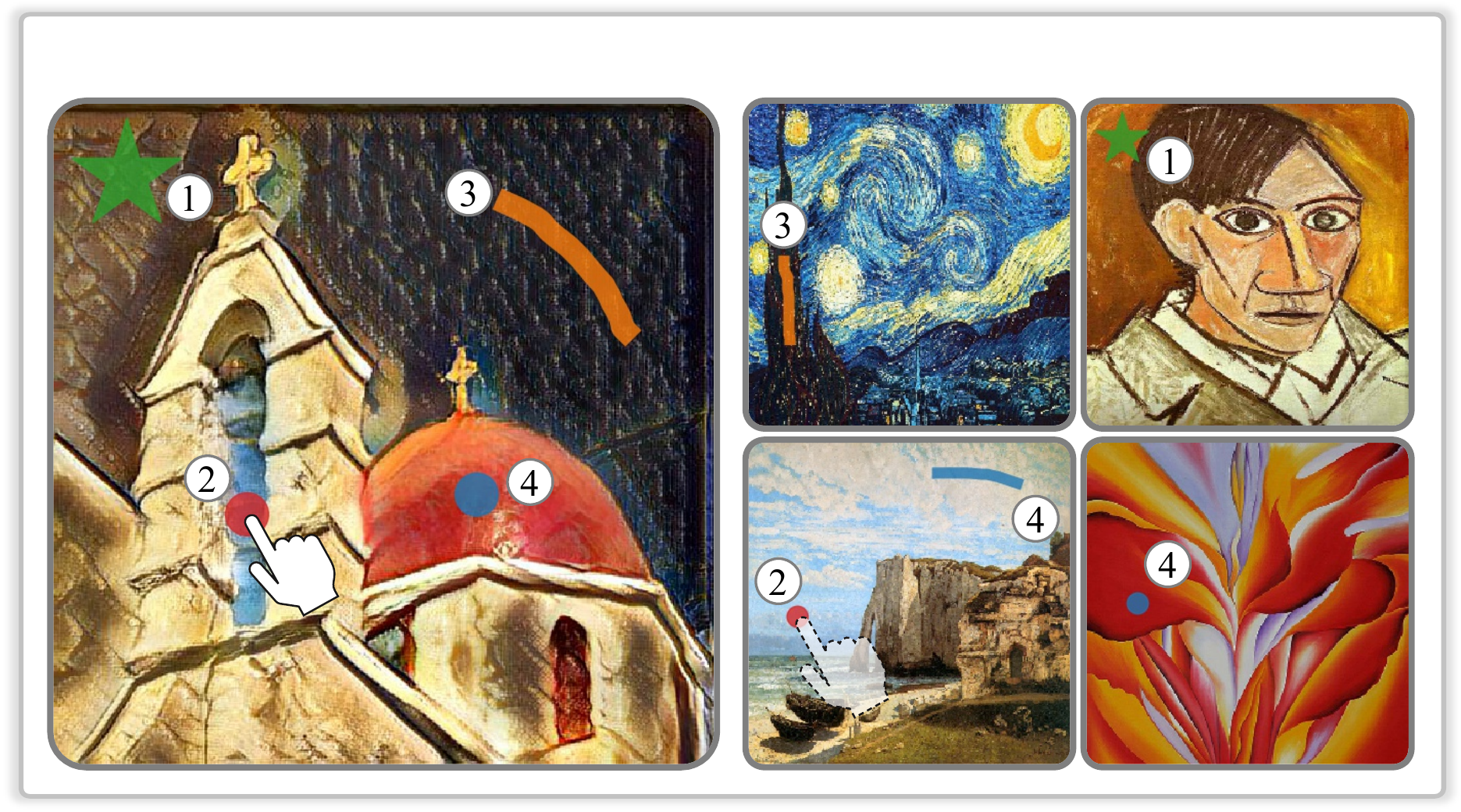} 
      \put(14.0,50.1){Content Canvas}
      \put(64.4,50.1){Style Palette}
    \end{overpic}
    \vspace{-15pt}
    \caption{User interface prototype. 
    The interaction mode can be global operation~\ding{192}, points~\ding{193}, and scribbles~\ding{194}.
    The operation \ding{195} mixes styles in different location and create a new style. 
    The interactions of the same color in the content and style images 
    represent the matching relationship between dipping and painting.
    } 
  \label{fig:ui}
  \vspace{-10pt}
  \end{figure}

%% file: main_parts/fig_abl-fluid.tex
\begin{figure*}[t]
	\centering
  \includegraphics[width=1.0\linewidth]{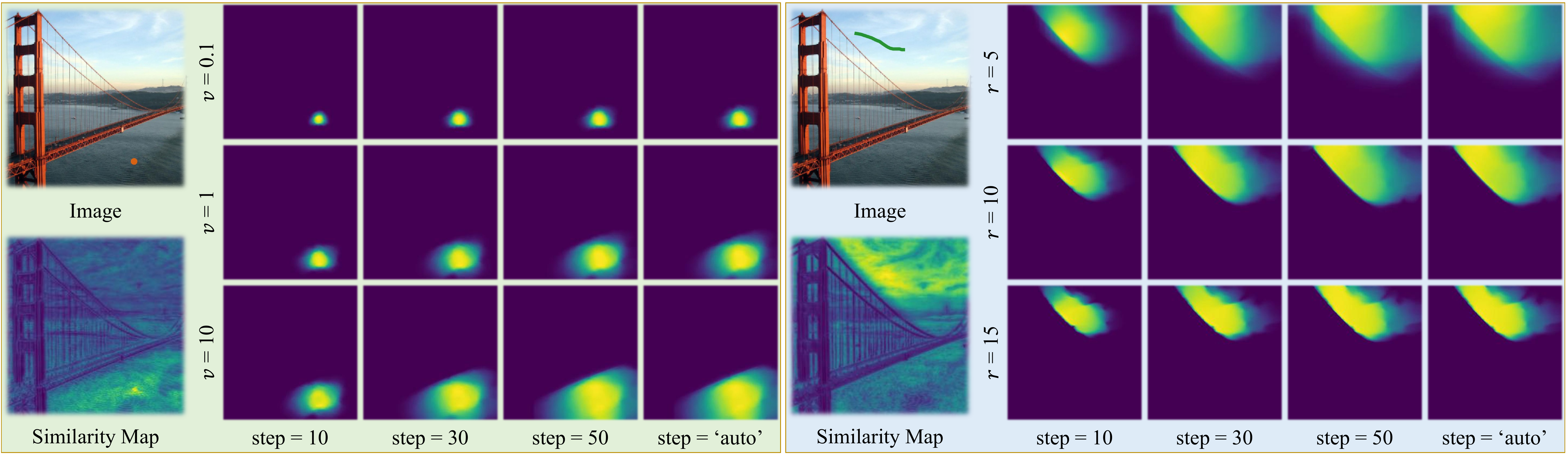}
  \vspace{-14pt}
  \vspace{-4pt}
	\caption{Comparison of the parameters in similarity-based fluid simulation algorithm. The $v$ and $r$ are the diffusion velocity and resistance coefficient, respectively.
  We show the animation with steps in 10, 30, 50,
  and the step from the automatic way.} 
  
\label{fig:abl-fluid}
\vspace{-5pt}
\vspace{-7pt}
\end{figure*}

%% file: main_parts/fig_iis.tex
\begin{figure*}[t]
	\centering
  \includegraphics[width=1.0\linewidth]{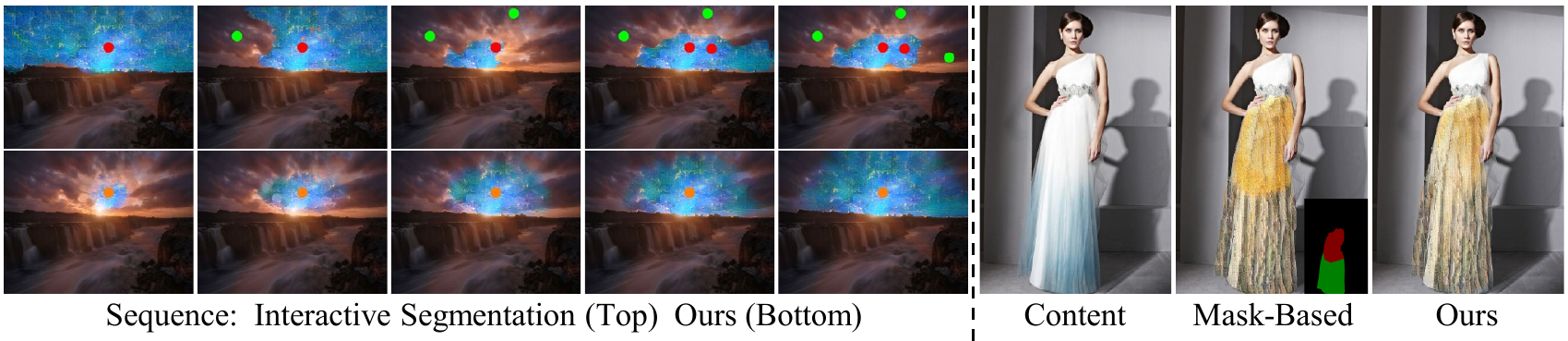}
  \vspace{-14pt}
  \vspace{-6pt}
	\caption{
    Comparison of visual effects between the mask-based transfer idea and our method.
    Left part: It shows an intuitive alternative idea which use interactive segmentation technology to segment local areas, and then stylize them. 
    We illustrate the visual effects and interaction logic of the alternative idea (the top row) and our method (the bottom row).
    We can see that our results are more controllable and in line with the logic of human drawing.
    Right part: It illustrates the transferred results by our method and the idea based on segmented masks.
    By our method,
    the transition of styles is smoother, and the fusion of different styles is more natural.
    } 
\label{fig:iis}
\vspace{-5pt}
\vspace{-9pt}
\end{figure*}

%% file: main_parts/fig_dip.tex
\begin{figure*}[!t]
	\centering
  \includegraphics[width=1.0\linewidth]{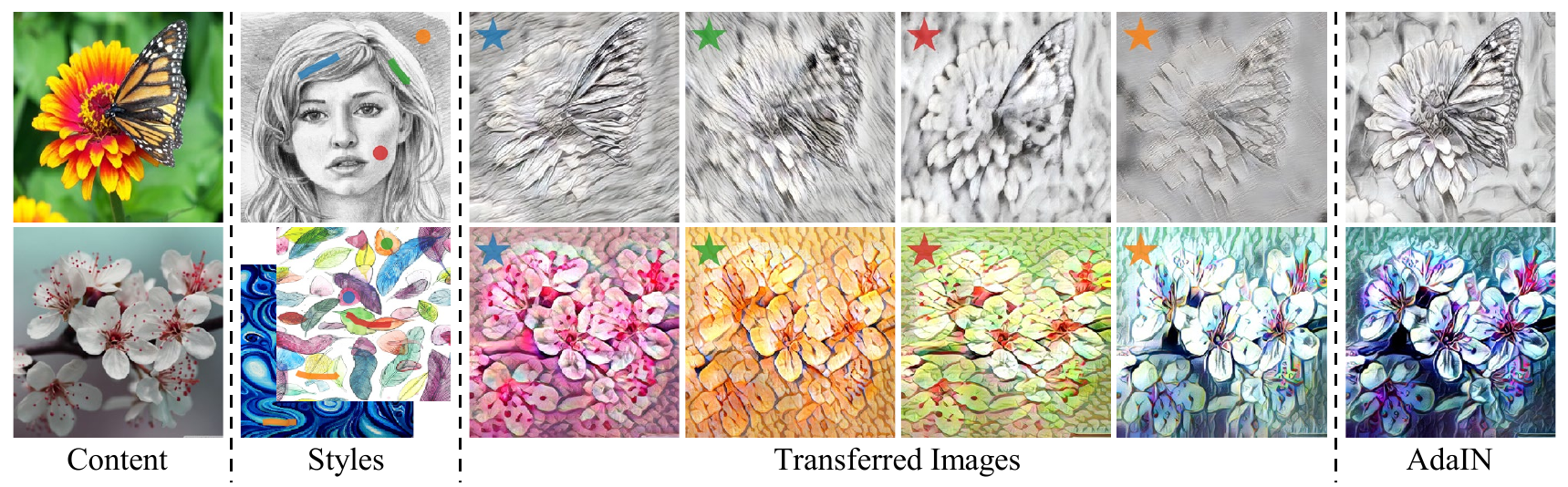}
  \vspace{-15pt}
  
	\caption{Dipping from different regions of style images.
    Compared to AdaIN,
    by our IST method,
    users can create diverse stylized results using limited style images.
    The first row shows the example of dipping in a single style.
    Users can create artworks of different texture directions and color concentrations according to their intention.
    The second row shows the dipping operation with merging styles in intra or inter images.
    The stylized results of our IST method can generate various color collocations according to user intention.} 
\label{fig:dip}
\vspace{-10pt}
\vspace{-0pt}
\end{figure*}

%% file: main_parts/fig_paint.tex
\begin{figure*}[t]
	\centering
  \includegraphics[width=1.0\linewidth]{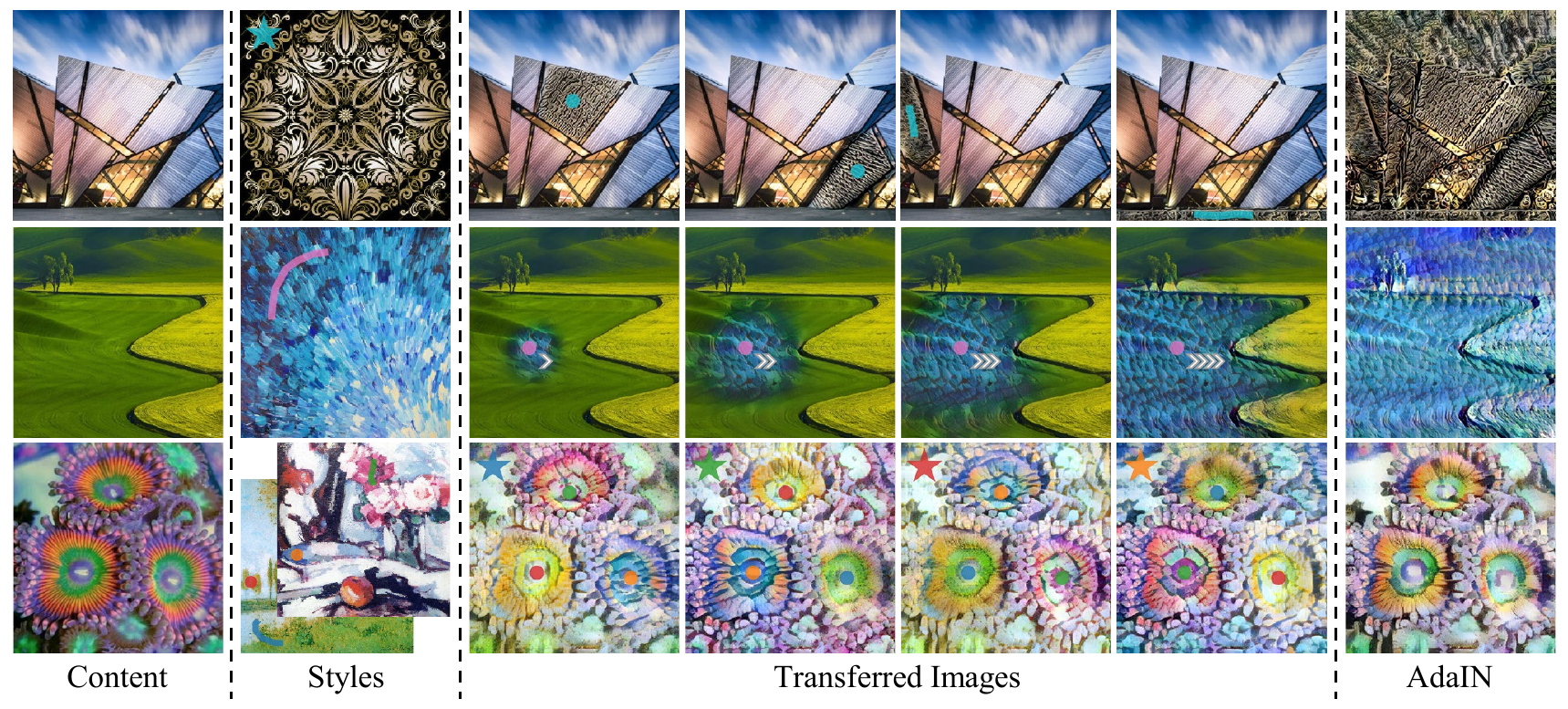}
  \vspace{-18pt}
	\caption{
        Painting into different regions of content images.
        The first row shows that the transferring in different areas of the building.
        The second row illustrates the style flow from the interaction position with the increase of holding time.
        In the third row,
        we dip four styles from style images and paint them to three flowers and the background and generate different results with different correspondence.
      }
\label{fig:paint}
\vspace{-14pt}
\end{figure*}

%% file: main_parts/fig_result.tex
\begin{figure}[t]
	\centering
  \includegraphics[width=1.0\linewidth]{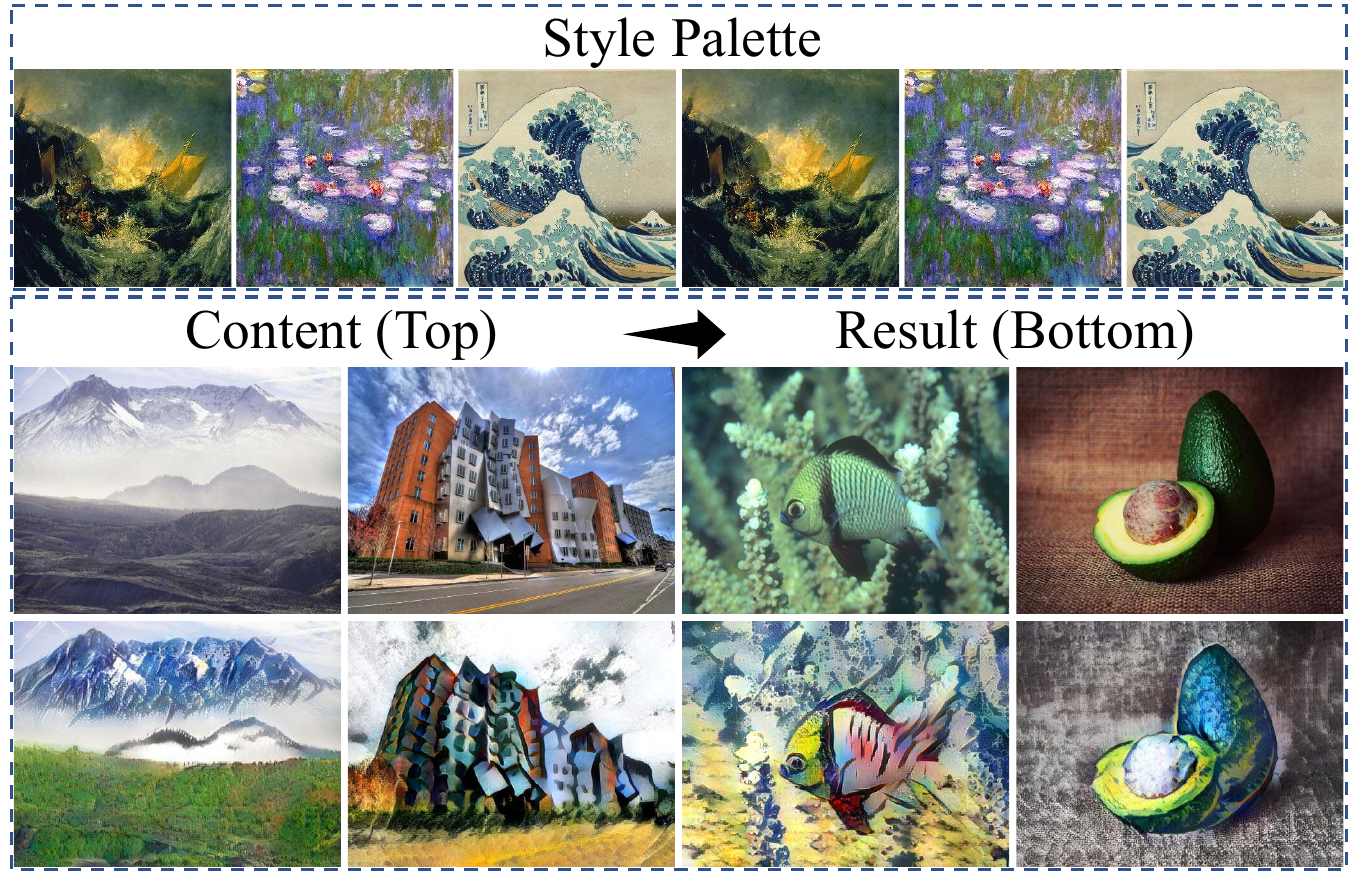}
  \vspace{-20pt}
	\caption{More artworks created by our IST method.}
\label{fig:result_all}
\vspace{-18pt}
\end{figure}